\documentclass{article}

\PassOptionsToPackage{numbers}{natbib}

\usepackage[preprint]{neurips_2025}

\usepackage[utf8]{inputenc} 
\usepackage[T1]{fontenc}    
\usepackage{hyperref}       
\usepackage{url}            
\usepackage{booktabs}       
\usepackage{amsfonts}       
\usepackage{nicefrac}       
\usepackage{microtype}      
\usepackage{xcolor}         

\usepackage{lipsum}		
\usepackage{graphicx}
\usepackage{doi}
\usepackage[utf8]{inputenc}
\usepackage{tcolorbox}
\usepackage{amsmath}
\usepackage{tabularx}
\usepackage{acronym}
\usepackage{makecell}

\acrodef{tom}[ToM]{Theory of Mind}
\acrodef{sft}[SFT]{Supervised Fine-Tuning}
\newcommand\boldgreen[1]{\textcolor{teal}{\textbf{#1}}}
\newcommand\boldred[1]{\textcolor{purple}{\textbf{#1}}}
\tcbset{
    mypromptbox/.style={
        colback=teal!5,     
        colframe=teal!70,    
        coltitle=white,      
        fonttitle=\bfseries, 
        boxrule=1mm,         
        arc=4mm,             
        left=2mm,            
        right=2mm,           
        top=2mm,             
        bottom=2mm,          
    }
}

\title{Do Theory of Mind Benchmarks Need Explicit Human-like Reasoning in Language Models?}

\author{ {Yi-Long~Lu}\thanks{Co-first authors. Code and data are available at \url{https://github.com/bigai-ai/ToM-RL.git}.},~\footnotemark[2] \quad
	{Chunhui Zhang}\footnotemark[1],~\footnotemark[2] \quad
        {Jiajun Song}, \quad
        {Lifeng Fan}, \quad
        {Wei Wang}\thanks{Corresponding authors} \quad \\
	State Key Laboratory of General Artificial Intelligence, BIGAI, Beijing, China\\
        \texttt{luyilong@pku.edu.cn, \{zhangchunhui, songjiajun, lifengfan, wangwei\}@bigai.ai}
}

\begin{document}
\maketitle
\begin{abstract}
  Theory of Mind (ToM), the ability to attribute mental states to others, is fundamental for human social intelligence and a critical capability for advanced Artificial Intelligence. Recent advancements in Large Language Models (LLMs) have shown promising performance on ToM benchmarks, raising the question: Do these benchmarks necessitate explicit human-like reasoning processes, or can models succeed through alternative strategies? We investigate this question empirically by applying Reinforcement Learning (RL) and Supervised Fine-Tuning (SFT) to LLMs of varying scales (0.5B to 7B parameters) and evaluating them across multiple ToM datasets. Our results reveal a scale-dependent impact of RL: while RL significantly improves accuracy and fosters high-quality, interpretable, and transferable belief-tracking reasoning in larger models (7B), it leads to "reasoning collapse" in smaller models ($\leq$3B), where high accuracy and generalization ability are achieved via drastically shortened, less meaningful responses. Surprisingly, further SFT achieves competitive and generalizable performance across these benchmarks, often matching or exceeding RL models in accuracy, despite not being explicitly trained to produce structured reasoning traces. These findings highlight a critical discrepancy between benchmark accuracy and the nature of learned reasoning. Our work suggests that current ToM benchmarks may be solvable without requiring the explicit, human-like simulation of mental states they were designed to probe. LLMs, particularly when scale is limited or training signals focus solely on output correctness, may leverage alternative rules effective for benchmark data structures. All the materials are available at {\it \href{https://sites.google.com/view/tom-reasoning}{tom-reasoning}}.
\end{abstract}

\section{Introduction}

\ac{tom}, the fundamental human ability to attribute mental states like beliefs, intentions, desires, and knowledge to oneself and others, is indispensable for effective social interaction, cooperation, and navigating complex social landscapes \citep{frith2005theory, leslie2004core}. As artificial intelligence systems become increasingly integrated into human environments, equipping them with a degree of \ac{tom} capability is crucial for fostering natural communication and collaboration \citep{cuzzolin2020knowing, williams2022supporting}. The remarkable emergence of large language models (LLMs), demonstrating sophisticated abilities across a wide array of cognitive tasks, has naturally led to significant interest in their potential for social cognition and mental state reasoning \citep{Kosinski_pnas2024, strachan2024testing}.

Existing research has developed various large-scale \ac{tom} benchmarks \citep{heHITOMBenchmarkEvaluating2023, le-etal-2019-revisiting, sclarExploreTheoryMind2024, shapira2023clever, gandhi2023understandingsocialreasoninglanguage}, often featuring multi-agent narratives designed to probe models' ability to track beliefs and predict actions based on agents' perspectives. While LLMs have shown promising results on these datasets \cite{sclarExploreTheoryMind2024, strachan2024testing}, a critical question arises: Do these benchmarks truly promote explicit human-like reasoning processes, such as simulating agents' mental states step-by-step, or can models achieve high performance by exploiting alternative strategies, potentially leveraging structural patterns inherent in the data? Answering this question is vital for accurately assessing AI's progress towards genuine social intelligence.

To empirically investigate this challenge, we employ two prominent post-training methodologies—Reinforcement Learning (RL) and Supervised Fine-Tuning (SFT)—on LLMs of different scales. RL, particularly with rule-based reward signals, has proven effective in enhancing structured reasoning in formal domains like mathematics and coding by reinforcing correct output and process adherence \citep{deepscaler2025, guo2025deepseek, xie2025logic}. While this suggests RL could potentially foster structured mental state reasoning \citep{gandhi2025cognitivebehaviorsenableselfimproving}, ToM involves context-sensitive social commonsense less amenable to rigid rules. SFT, on the other hand, directly optimizes models to reproduce desired outputs from provided examples. By comparing the performance and, crucially, the nature of the reasoning elicited by RL and SFT across models of varying capacities, we can gain insight into the strategies LLMs employ to solve current ToM tasks and whether these strategies align with explicit belief-tracking.

Our study yields several key findings that shed light on the nature of reasoning on current ToM benchmarks. Firstly, we find that while RL significantly boosts accuracy across models of different sizes, its impact on reasoning quality is scale-dependent. In larger models (7B), RL induces high-quality, interpretable, and transferable belief-tracking behaviors. However, in smaller models ($\leq$3B), RL leads to reasoning collapse: models achieve high accuracy and generalization but produce drastically shortened, less meaningful responses, suggesting they rely on implicit rather than explicit structured reasoning. Secondly, and perhaps most strikingly, we demonstrate that SFT alone achieves competitive and generalizable performance on these benchmarks, often matching or exceeding RL models in accuracy, despite not being explicitly optimized for the reasoning process.

These results highlight a critical discrepancy between achieving high scores on current ToM benchmarks and demonstrating explicit human-like reasoning. Our work suggests that existing benchmarks may be solvable without requiring genuine, step-by-step simulation of mental states. LLMs, especially those with limited capacity or trained with output-focused signals, may internalize alternative rules or patterns that are effective for the specific structures found in benchmark datasets.

This paper makes the following key contributions:
\begin{itemize}
    \item We achieve state-of-the-art performance and reveal a scale-dependent effect of RL on LLM in \ac{tom}: it promotes explicit reasoning in 7B models but leads to reasoning collapse (high accuracy without meaningful reasoning) in smaller models, revealing a crucial mismatch between performance and reasoning quality.
    \item We show that SFT achieves competitive and generalizable performance on current \ac{tom} benchmarks, providing empirical evidence that these datasets may not require explicit human-like mental state reasoning.
    \item We highlight the need for future \ac{tom} evaluation methods and benchmarks that assess not just answer accuracy but the depth and nature of underlying reasoning.
\end{itemize}

\section{Background and Related Work}
\subsection{Theory of Mind in Psychology}
\ac{tom} is a core concept in psychology referring to the capacity to understand and attribute mental states—such as beliefs, desires, intentions, and knowledge—to oneself and others. This ability is crucial for predicting behavior, navigating social interactions, and understanding communication. A classic method for assessing \ac{tom}, particularly in developmental psychology, is the false-belief task. These tasks probe whether an individual can understand that another person may hold a belief that is contrary to reality, and predict their actions based on that false belief. 

Consider a classic \ac{tom} scenario: \textit{Sally places her marble in a basket and leaves; Anne moves it to a box. Where will Sally search for her marble when she returns?} The correct answer depends not on physical laws, but on understanding Sally's false belief. This kind of ability is critical to human social interactions~\citep{Kosinski_pnas2024, strachan2024testing}. The complexity of these tasks can be increased by nesting beliefs, moving from first-order (e.g., "Where does Anne think the marble is?") to higher-orders (e.g., "Where does Anne think Sally thinks the marble is?").

\subsection{ToM in Artificial Intelligence}
Early AI approaches to ToM involved neural networks designed to track agents' beliefs in controlled environments\citep{grant2017can, nematzadeh-etal-2018-evaluating, le-etal-2019-revisiting}. With the advent of large language models (LLMs), research has shifted to exploring whether these models can inherently perform ToM-like reasoning. Recent studies have systematically examined LLMs' performance on various ToM tasks, including false-belief scenarios \citep{Kosinski_pnas2024, strachan2024testing, ullman2023large, shapira2023clever}. Benchmarks such as Hi-ToM \citep{heHITOMBenchmarkEvaluating2023}, Explore-ToM \citep{sclarExploreTheoryMind2024}, and ToMi \citep{le-etal-2019-revisiting} have been developed to assess LLMs on tasks ranging from simple first-order beliefs to complex higher-order reasoning. While large models like GPT-4 have shown proficiency in standard false-belief tasks \citep{Kosinski_pnas2024}, performance tends to degrade significantly on higher-order problems \citep{heHITOMBenchmarkEvaluating2023}. Recent efforts have also focused on building larger ToM datasets and fine-tuning smaller LLMs to improve social cognition reasoning in more efficient models \citep{sclarExploreTheoryMind2024}. 

\subsection{Rule-based reinforcement learning in LLMs}
Rule-based reinforcement learning (RL) has proven effective in enhancing large language models (LLMs) beyond standard supervised fine-tuning. This approach uses structured reward signals to guide model behavior without requiring explicit step-by-step supervision. Notably, it enables the emergence of reasoning-like behavior through relatively simple feedback mechanisms. Recent work has shown that such rewards can encourage models to internalize structured thinking. For instance, DeepSeek-R1 \citep{guo2025deepseek} demonstrated that answer-level rewards led to gradual increases in both response length and accuracy, suggesting the emergence of reasoning dynamics. Logic-RL \citep{xie2025logic} applied format-constrained rewards to synthetic logic puzzles, training a 7B model that learned to reflect and verify its answers—skills that transferred to real-world math benchmarks such as AIME. Similarly, SWE-RL \citep{weiSWERLAdvancingLLM2025} applied similar principles to software engineering, achieving state-of-the-art performance on coding tasks and exhibiting cross-domain generalization. These successes highlight RL's potential to activate latent reasoning skills through reward design. However, prior successes have largely occurred in formal domains, where rules and ground truths are well-defined. It remains an open question whether such methods generalize to social reasoning tasks, which require interpreting mental states and hidden commonsense. The extent to which rule-based RL can elicit human-like mental state inference in LLMs remains underexplored.


\section{Experimental Methods}
\subsection{Theory of Mind Benchmarks}
To assess whether RL enables LLMs to acquire explicit mind reasoning, we curated a diverse set of \ac{tom} benchmarks. Guided by prior work\citep{xie2025logic}, we prioritized datasets with clean structure, well-defined and verifiable answers, and non-trivial reasoning requirements. These benchmarks span varying levels of difficulty and include diverse question formats and narrative styles. This design allows for controlled analysis across model sizes and training methods, isolating the impact of reinforcement learning on belief modeling.

\subsubsection{Training data}
We generated an initial Hi-ToM-style dataset of 3,000 samples using the data generation script from \citet{heHITOMBenchmarkEvaluating2023}. This dataset was evenly distributed across ToM orders: 0 (factual questions), 1, 2, 3, and 4, with 600 samples per level. To assess out-of-distribution (OOD) performance, \textbf{all 600 fourth-order samples were reserved for testing}. From the remaining 2,400 samples, we randomly selected 2,000 for training and 400 for validation. 

To further enhance the diversity of the training set, we incorporated 1,200 samples from ExploreToM's dataset \citep{sclarExploreTheoryMind2024}, covering a range of story types. We excluded all binary-response questions (e.g., "Yes" or "No") from training, as they were highly biased in the ExploreToM dataset \footnote{Our preliminary results indicated that such biases significantly impaired the model’s ability to learn correct reasoning patterns.}. As a result, the final training dataset comprised 3,200 samples across 900 unique stories. During training, we relied solely on the structured story formats provided by ExploreToM.

\subsubsection{Evaluation data}
For evaluation, we used multiple datasets to assess both in-distribution performance and generalization capabilities. In addition to the 600 4th-order OOD ToM samples generated by the script, we also incorporated the original Hi-ToM dataset \citep{heHITOMBenchmarkEvaluating2023}, comprising 600 examples across ToM orders 0 to 4. We also evaluated model performance on the ToMi dataset \citep{le-etal-2019-revisiting}, which covers unexpected transfers as well as first- and second-order false belief tasks. To further examine generalization, we tested on the ExploreToM benchmark \citep{sclarExploreTheoryMind2024}, using 1,066 questions with raw story structures. We further tested robustness to surface variation using an infilled version, where stories were enriched with contextual details and rewritten into natural narratives. (Appendix Table~\ref{tab:datasets}).

The final evaluation has 5 different datasets: the original Hi-ToM, 4th-order ToM (OOD), ToMi dataset, 1066 ExploreToM dataset with raw story structures, and infilled story version.

\subsection{Models}
\paragraph{Trainning Models} To assess whether reinforcement learning (RL) enhances \ac{tom} reasoning in small-scale LLMs and to examine how model size influences both reasoning processes and RL effectiveness, we trained a range of Qwen 2.5 instruct models. Prior research suggests that Qwen models naturally exhibit reasoning behaviors that facilitate self-improvement during RL training \citep{gandhi2025cognitivebehaviorsenableselfimproving}. Building on these findings, we trained five models with RL and SFT: Qwen2.5 instruct version at 0.5B, 1.5B, 3B, 7B, and 7B-1M\footnote{7B-1M stands for 1M-token context version of Qwen 2.5.} \citep{qwenQwen25TechnicalReport2025, yangQwen251MTechnicalReport2025}, spanning a range from small to moderately-sized.

\paragraph{Baseline LLMs} Besides Qwen models, for baseline reference, we also included GPT-4o-mini, GPT-4o, as well as the DeepSeek-v3 model \citep{deepseek-aiDeepSeekV3TechnicalReport2024}. 
\subsection{Training Methodologies}
\subsubsection{Rule-based reinforcement learning}
The original Hi-ToM dataset includes a list of assumptions within its prompts. We streamlined these by retaining only two key assumptions in our system prompts during training (see our system prompt \ref{tcolorbox:sys_prompt}).

\paragraph{Format reward}
Following prior works \citep{deepseek-aiDeepSeekV3TechnicalReport2024, xie2025logic, zhou2025r1}, we implemented a format-based reward to encourage structured reasoning. Specifically, we required models to enclose their reasoning process within \texttt{<think></think>} tags \footnote{To facilitate learning process, we manually add \texttt{<think>} at the end of chat template.} and their final answers within \texttt{<answer></answer>} tags. 

\begin{equation}\label{eq:format}
R_{\text{format}} =
    \begin{cases} 
        1, & \text{if the format is correct} \\
        -1, & \text{if the format is incorrect}
    \end{cases}
\end{equation}

\paragraph{Answer reward} 
To check and reinforce the correct answer, we used a regular expression-based approach to evaluate answer accuracy. This method accounts for variations in response style, as the standard answers in the dataset are typically single words or phrases, whereas model-generated responses often take the form of complete sentences. 
\begin{equation}
    R_{\text{answer}} =
    \begin{cases} 
        2, & \text{if the format is correct and the answer matches the ground truth} \\
        -2, & \text{if the format is incorrect or the answer cannot match the ground truth}
    \end{cases}
\end{equation}
We summed $R_{\text{format}}$ and $R_{\text{answer}}$ to obtain the final reward signal for training.

\paragraph{RL algorithm}

We adopt GRPO, the reinforcement learning baseline algorithm proposed in DeepSeekMath \citep{shaoDeepSeekMathPushingLimits2024}, as our RL foundation. GRPO has already demonstrated strong potential in DeepSeek-R1~\citep{guo2025deepseek} and has recently been shown to be stable and reproducible across a series of follow-up works~\citep{deepscaler2025,weiSWERLAdvancingLLM2025}. To reduce the overhead of training an additional value function as in PPO~\citep{schulmanProximalPolicyOptimization2017}, GRPO instead uses the average reward of sampled responses as a baseline for computing advantages. 

The training process was conducted using the veRL framework \citep{sheng2024hybridflow}. We set the training batch size to 8 and the number of rollout generations to 16. A consistent learning rate of 5e-7 and a temperature parameter of 1 were used to encourage greater model exploration. The model was trained for 2 epochs on 3,200 samples, resulting in a total of 800 optimization steps. All experiments were performed on 4 A100 (80GB) GPUs.

\begin{table}[t]
    \centering
    \caption{Model performance across different \ac{tom} evaluation benchmarks.}
    \begin{tabular}{lccccc}
    \toprule
    Model & \makecell{4th-order\\ToM} & Hi-ToM & ToMi & \makecell{ExploreToM\\(Raw)} & \makecell{ExploreToM\\(Infilled)} \\
    \midrule
    GPT-4o & 46.17\% & 69.00\% & 61.96\% & 67.35\% & 62.48\% \\
    GPT-4o-mini & 30.50\% & 58.50\% & \textbf{70.64\%} & \textbf{69.32\%} & \textbf{66.04\%} \\
    DeepSeek-v3 & \textbf{58.67\%} & \textbf{70.17\%} & 57.17\% & 65.01\% & 64.73\% \\
    \midrule
    Qwen2.5-0.5B-Instruct & 23.83\% & 30.33\% & 29.38\% & 60.51\% & 54.97\% \\
    Qwen2.5-1.5B-Instruct & 25.17\% & 40.67\% & 54.12\% & 54.78\% & 43.53\% \\
    Qwen2.5-3B-Instruct & 27.50\% & 39.17\% & 47.78\% & 47.37\% & 49.81\% \\
    Qwen2.5-7B-Instruct & 28.83\% & 52.17\% & 54.65\% & 59.38\% & 45.40\% \\
    Qwen2.5-7B-Instruct-1M & 17.83\% & 40.67\% & 54.85\% & 37.34\% & 41.46\% \\
    \midrule
    Qwen2.5-0.5B-Instruct (RL) & 85.83\% & 70.83\% & 54.25\% & \textbf{93.34\%} & 72.61\% \\
    Qwen2.5-1.5B-Instruct (RL) & 89.33\% & 79.17\% & 75.89\% & 90.06\% & 70.73\% \\
    Qwen2.5-3B-Instruct (RL) & 88.17\% & 81.17\% & 80.18\% & 93.43\% & \textbf{78.14\%} \\
    Qwen2.5-7B-Instruct (RL) & 82.83\% & 83.33\% & 73.99\% & 91.65\% & 74.77\% \\
    Qwen2.5-7B-Instruct-1M (RL) & \textbf{94.50\%} & \textbf{84.50\%} &\textbf{81.08\%} & 92.31\% & 77.20\% \\
    \midrule
    Qwen2.5-0.5B-Instruct (SFT) & 88.17\% & 81.00\% & 77.79\% & 89.68\% & 69.89\% \\
    Qwen2.5-1.5B-Instruct (SFT) & 86.17\% & 80.50\% & 76.33\% & 93.53\% & 74.67\% \\
    Qwen2.5-3B-Instruct (SFT) & 92.67\% & 87.00\% & 79.55\% & 95.78\% & 74.95\% \\
    Qwen2.5-7B-Instruct (SFT) & \textbf{94.00\%} & \textbf{87.33\%} & 80.85\% & \textbf{95.97\%} & \textbf{77.95\%} \\
    Qwen2.5-7B-Instruct-1M (SFT) & 93.67\% & 86.50\% & \textbf{81.10\%} & 95.12\% & 75.61\% \\
    \bottomrule
    \end{tabular}
    \label{tab:tom_results}
\end{table}
\subsubsection{Supervised Fine-Tuning}
SFT adapts pre-trained LLMs to specific tasks via input-output training. It provides a baseline for evaluating whether complex methods like RL offer real gains. On ToM tasks, the comparisons between SFT and RL help assess if models rely on true reasoning or surface-level patterns. We fine-tuned Qwen models using the veRL framework with full-parameter updates on ground truth answers. Training used a batch size of 256 and AdamW (lr=1e-5) on the same 3,200 samples.

\subsection{Evaluation metrics}
All model outputs are scored via regex-based answer matching to compute the accuracy. Each input follows a standard format: 
\begin{verbatim}
Story: {story}\n Question: {question}
\end{verbatim}
Baseline models use Chain-of-Thought (CoT) prompting with thinking in JSON responses (See Appendix~\ref{tcolorbox:cot_prompt}). RL models use DeepSeek-R1-style prompts (Appendix~\ref{tcolorbox:sys_prompt}). The SFT model is evaluated with default prompting.

\paragraph{Thinking quality}
To assess the reasoning quality for RL-trained models, we use GPT-4.1 to rate model-generated reasoning on 600 fourth-order ToM questions, which are considered challenging even for humans. Two criteria are scored on a 0–10 scale: \textbf{logical coherence} (is the reasoning internally consistent?) and \textbf{factual alignment} (does it reflect the story’s content accurately?). The final thinking quality score is the average of these two ratings, normalized to the [0, 1] range. Prompt details are in the Appendix \ref{tcolorbox:think_prompt}.

\paragraph{Knowledge transfer performance} To further examine the quality of model-generated reasoning, we use the thinking content produced by the 7B-1M and 3B models on fourth-order ToM questions as input to GPT-4o-mini, prompted with CoT. We evaluate GPT-4o-mini’s answer accuracy using both pre- and post-RL thinking traces. Since the final sentence often summarizes the reasoning and contains the answer, we also removed this conclusion before passing the content, ensuring that GPT-4o-mini relies only on the reasoning process itself.

\section{Results}

\subsection{RL Performance on ToM Benchmarks}
\paragraph{Reinforcement learning improves \ac{tom} performance} 
Table \ref{tab:tom_results} presents the model perfermance across different datasets. Compared to baseline models, RL led to a performance increase of approximately 40\% on the Hi-ToM benchmark. The 7B-1M model achieved an accuracy of 84.50\%, outperforming GPT-4o (69.00\%) and DeepSeek-v3 (70.17\%). Notably, despite being trained only on ToM questions up to the third order, our model successfully generalized to fourth-order \ac{tom} questions—a level of reasoning that proved extremely challenging for baseline models and humans. Interestingly, the performance of 0.5B model also increased from 23.83\% to 85.83\% in 4th-order \ac{tom}.

On the ToMi and ExploreToM datasets, RL-trained models also surpassed their baseline counterparts, even when tested on longer, more detailed input stories (ExploreToM with infilled stories). Given that our models are significantly smaller than GPT-4o and DeepSeek-v3, these results highlight the potential of RL in enhancing ToM capabilities in small LLMs. 

\paragraph{RL does not impair performance in other domain} To assess whether training on our \ac{tom} dataset impacted performance in other domains, we also evaluated the model on 1319 GSM8K math test samples\footnote{We found that different prompts significantly affected Qwen's mathematical performance. To maximize the baseline model's performance, we adopted the following prompt: "Let's think step by step and output the final answer within \textbackslash  \textbackslash boxed\{\}."}. Unlike previous study\citep{xie2025logic}, evaluation results (Appendix Table \ref{tab:gsm8k_results}) indicated that reinforcement learning did not improve (but also did not impair) the model's performance on entirely out-of-distribution reasoning tasks like math.

\subsection{Qualitative Analysis of Reasoning Processes}
\begin{figure*}[t]
    \centering
    \includegraphics[width=1\textwidth]{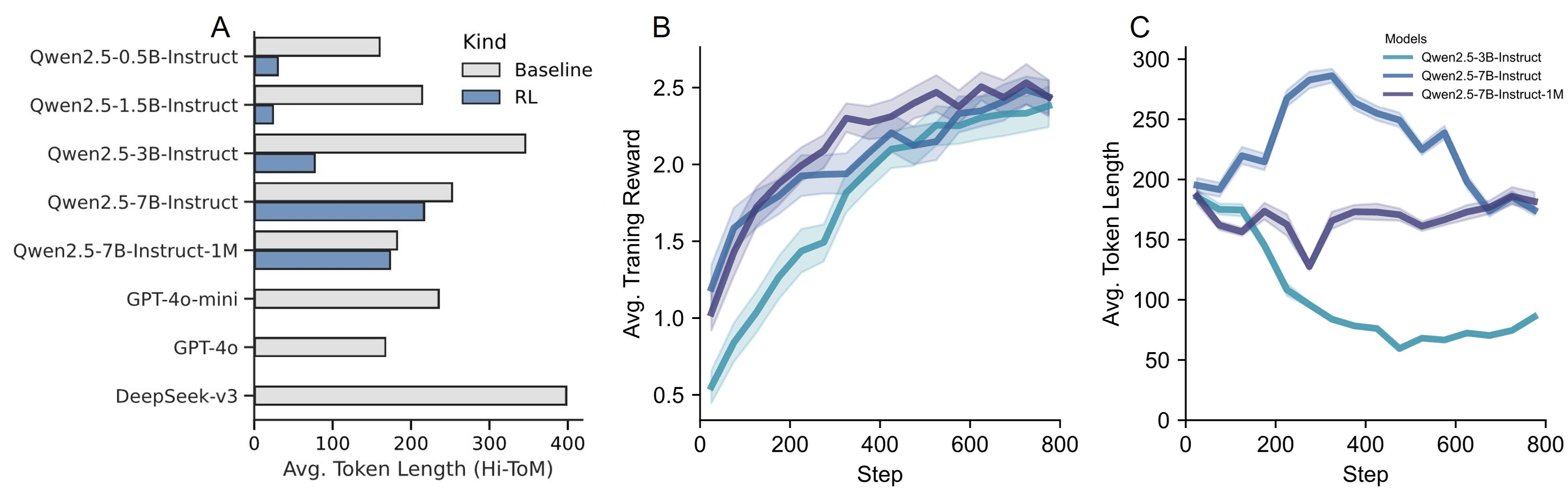}
    \caption{Model response length and training dynamics. (A) Average response length on Hi-ToM, where models smaller than 3B collapsed their responses after RL training. (B) Training scores continued to rise throughout training (only 3B, 7B, and 7B-1M models are shown for visualization). (C) Response length dynamics during training: the 3B model (light blue) showed a decreasing trend, while the 7B models maintained longer responses. The color bands stand for 90\% CI.}
    \label{fig:response}
\end{figure*}

\paragraph{Reasoning collapse in smaller models ($\leq$ 3B)}
To assess whether models acquired genuine social reasoning or exploited dataset artifacts, we tracked response length before and after reinforcement learning (RL). Prior studies suggest that response length tends to increase with improved performance \citep{deepscaler2025, deepseek-aiDeepSeekV3TechnicalReport2024, xie2025logic}. 
However, as shown in Figure~\ref{fig:response}A, we observed the opposite in smaller models ($\leq$3B): response lengths dropped significantly after RL. The 1.5B model, for example, reduced its average length by 88.4\% while accuracy rose from 40.67\% to 79.17\%. In contrast, the 7B model maintained consistent response length. Figures~\ref{fig:response}B\&C further show that, despite steady reward increases, smaller models continued shortening their outputs.

\paragraph{Larger models develop structured reasoning}
We then examined the reasoning processes generated by models of varying scales. In general, the 7B-1M model explicitly tracked agent actions and belief states, often distinguishing between shared and private knowledge. Despite occasional errors, it maintained coherent mental state representations across multi-step scenarios, suggesting a structured reasoning strategy. In contrast, the 3B model relied on shallow reasoning, often approximating belief propagation without maintaining precise mental state representations. The 1.5B and 0.5B models frequently repeated the final answer or mimicked prompt structure, producing minimal intermediate reasoning (see Appendix~\ref{tcolorbox:response_example} \& \ref{tcolorbox:response_example_wrong} for correct and wrong reasoning examples).


\begin{figure*}[t]
    \centering
    \includegraphics[width=1\textwidth]{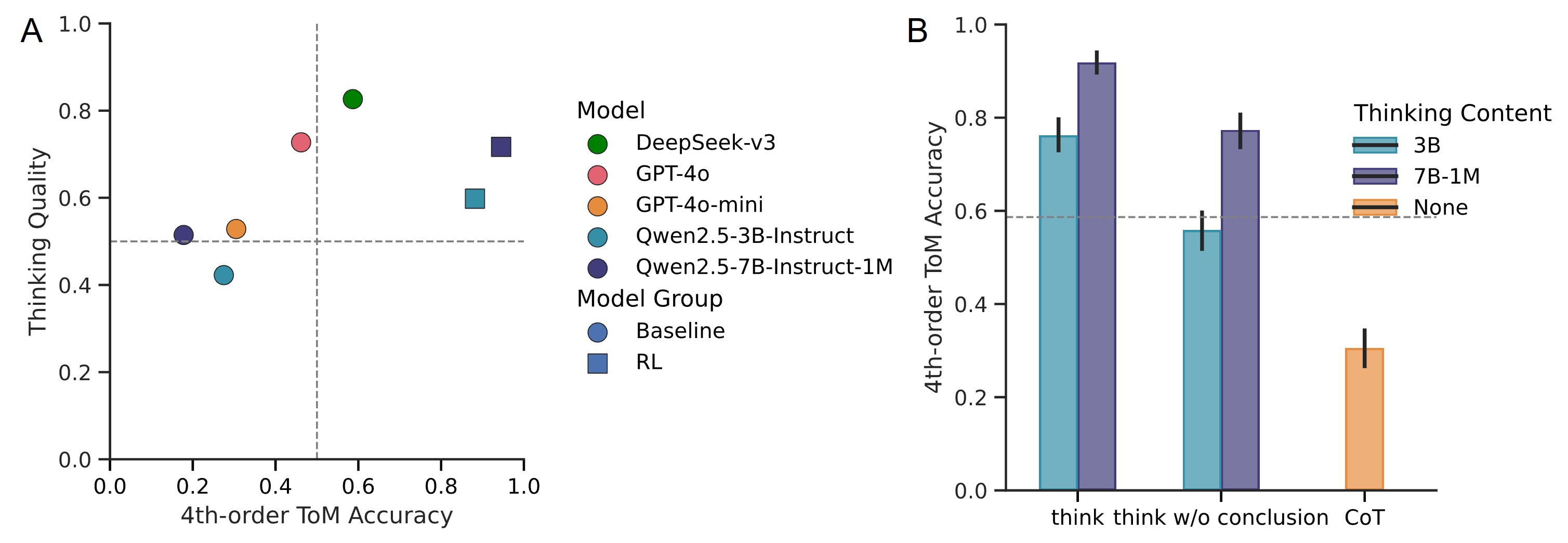}
    \caption{Analysis of model reasoning on fourth-order ToM tasks. (A) ToM accuracy versus thinking quality. The 7B-1M model achieves the highest accuracy with good reasoning quality. (B) Teaching GPT-4o-mini using model-generated reasoning. The 7B-1M's reasoning helps GPT-4o-mini surpass DeepSeek-v3 (gray line), even when conclusions are removed from the input. Error bars stand for 90\% CI.}
    \label{fig:think}
\end{figure*}

We assessed reasoning quality on fourth-order ToM examples using GPT-4.1, which scored each response based on logical coherence and factual alignment. As shown in Figure~\ref{fig:think}A, DeepSeek-v3 reached the highest reasoning quality (0.826), followed by GPT-4o (0.727). Among our models, the 7B-1M model achieved lower reasoning quality than DeepSeek-v3 and GPT-4o but outperformed GPT-4o-mini (0.528), improving from 0.515 to 0.717 after RL.  Notably, it attained the highest accuracy overall (94.5\%). The 3B model also showed gains in reasoning quality, increasing from 0.423 to 0.598 after RL, despite generating shorter responses (see Appendix Table~\ref{tab:model-eval} for detailed scores).

To test whether these reasoning traces carry transferable knowledge, we passed them as demonstrations to GPT-4o-mini and measured its performance. If the traces encode correct beliefs and inferences, they should boost downstream accuracy. Results in Figure~\ref{fig:think}B confirm this. Using 7B-1M traces improved GPT-4o-mini's accuracy from 30.5\% to 91.83\%. Even after removing final conclusions from the traces, performance remained high (77.33\%). The 3B model’s traces provided moderate gains (76.17\% full, 55.83\% without conclusion), suggesting less consistent internal structure.

These results indicated that RL can teach larger models to construct reusable reasoning strategies. Smaller models instead optimize toward output-level accuracy using implicit rules or shallow reasoning. The sharp contrast also suggested a potential limitation in current ToM benchmarks, which may reward correct answers without enforcing structured mental state inference \citep{heHITOMBenchmarkEvaluating2023, sclarExploreTheoryMind2024, le-etal-2019-revisiting}.

\vspace{3em}
\subsection{SFT on ToM Benchmarks}
To isolate the effects of reinforcement learning, we compared SFT-only models with their RL-finetuned counterparts across multiple ToM benchmarks. Table~\ref{tab:tom_results} summarizes results for all model sizes on Hi-ToM, 4th-order ToM, ToMi, and ExploreToM (standard and infilled variants).

Surprisingly, SFT alone achieved near-identical performance to RL in several cases. For the 7B-1M model, SFT even outperformed RL slightly on Hi-ToM (86.5\% vs 84.5\%). This trend persisted across other model sizes. The SFT models successfully generalized to 4-th order ToM and infilled ToM stories, with the 0.5B SFT model reaching 88.17\% and 69.89\% respectively. Overall, SFT offers a surprisingly strong baseline, often matching RL on ToM tasks with different stories and question orders. These results still hold when we reduce the trainning data to one-quarter (800 samples, see Appendix Table \ref{tab:sft_results}). However, further evaluation on GSM8k showed that, unlike RL, SFT significantly impairs math performance, with scores dropping close to zero.


\section{Discussion}
Our empirical investigation reveals a complex interplay between model scale, training methodology, and the acquisition of Theory of Mind capabilities in Large Language Models. The findings highlight both the potential of reinforcement learning and the limitations of current evaluation paradigms for social reasoning tasks.

\paragraph{RL's success in inducing explicit reasoning in larger models}
We demonstrated that RL significantly boosts performance on a range of \ac{tom} benchmarks across all tested Qwen models. Larger models (7B-1M) trained with RL not only achieved state-of-the-art accuracy on complex 4th-order \ac{tom} tasks, surpassing significantly larger models like GPT-4o and DeepSeek-v3 on certain benchmarks, but also developed some interpretable and transferable reasoning processes. Our analysis using GPT-4.1 for thinking quality assessment and the knowledge transfer experiment with GPT-4o-mini provides converging evidence that RL can induce more structured belief tracking in models with sufficient capacity. This aligns with prior work showing RL's ability to foster emergent reasoning in domains like mathematics and coding \citep{guo2025deepseek, xie2025logic, weiSWERLAdvancingLLM2025}, suggesting that sophisticated reasoning skills can be reinforced through task-specific rewards.

\paragraph{Reasoning collapse: the pitfall of RL in smaller models}
The beneficial effect of RL on reasoning quality was not uniform across all model sizes. The reasoning collapse emerged in smaller models ($\leq$3B). Despite achieving substantial accuracy gains comparable to the larger models on many benchmarks after RL training, these models failed to generate interpretable, structured reasoning traces. Instead, they appeared to rely on shorter, potentially memorized patterns or rules optimized directly for the final answer, rather than explicitly tracking agents' mental states. This phenomenon underscores a crucial mismatch between achieving high accuracy on benchmark questions and possessing genuine, human-like reasoning capabilities. The simple rule-based rewards, while effective at optimizing for correctness, may inadvertently encourage shortcut learning in models with limited capacity relative to the complexity of the underlying task logic. 

\paragraph{Effectiveness of SFT and implications for benchmark validity}
The surprisingly strong performance of models trained solely with SFT further supports the notion that existing \ac{tom} benchmarks, such as Hi-ToM and ExploreToM, while valuable, may not be sufficiently challenging to exclusively probe deep mental state inference. SFT models achieved accuracy comparable to or even slightly better than their RL-trained counterparts on several datasets, including those designed to test generalization (4th-order ToM, infilled stories). One possible explanation is that the datasets might contain exploitable patterns, such as surface-level correlations between narrative elements and answers, possibly introduced by templated generation. For example, in ExploreToM\citep{sclarExploreTheoryMind2024}, 22\% of questions have “yes” as the correct answer, while only 4\% are “no,” introducing a strong prior. Additionally, general pretraining might equip models with reasoning skills that can be activated by SFT. The generalization seen with SFT also implies that simply increasing the complexity or naturalism of the stories (as in infilled ExploreToM) might not be enough to overcome potential implicit strategies if the underlying logical structure remains predictable.

Our findings have important implications for the development and evaluation of social intelligence in AI. First, achieving high accuracy on current \ac{tom} benchmarks should not be taken as definitive proof of human-level understanding of mental states. Evaluation methods must evolve to directly assess the quality and robustness of the reasoning process, perhaps through counterfactual questioning, requiring explanations for belief states, or using adversarial generation techniques specifically targeting reasoning chains rather than just final answers \citep{shapira2023clever,ullman2023large}. Second, training methodologies, including RL, need to be carefully designed and potentially adapted based on model scale. For smaller models, simpler rewards focused solely on output correctness might inadvertently lead to non-interpretable, brittle solutions. Future work could explore curriculum learning, auxiliary losses that encourage intermediate state tracking, or reward functions derived from desired reasoning properties (e.g., consistency of belief attribution). 

\subsection{Limitations}
Our study has several limitations. In our experiments, we did not evaluate recent reasoning models such as o3 and DeepSeek-R1. However, our goal was not to achieve state-of-the-art performance but to assess whether current ToM datasets can elicit explicit mind reasoning via RL. The evaluation of thinking quality relies on GPT-4.1, which is itself an LLM, and its judgment may not perfectly capture human assessment of reasoning or be free from its own biases. The knowledge transfer experiment using GPT-4o-mini similarly depends on the capabilities of another LLM. What exactly is being transferred needs further investigation. While these methods offer valuable proxies for reasoning structure, human evaluation would provide a stronger gold standard. We focused on a specific family of models (Qwen 2.5) and a particular RL setup (GRPO with simple rewards); results may vary with different architectures, model families, or RL algorithms and reward designs. Furthermore, our training data was generated from synthetic sources; evaluating performance on more naturalistic or human-annotated \ac{tom} scenarios would be a valuable direction for future work.

\section{Conclusion}
In this paper, we investigated the effectiveness of reinforcement learning for enhancing Large Language Models' ability to reason about mental states, a core aspect of Theory of Mind. We trained Qwen 2.5 models of varying sizes using rule-based RL on synthetic \ac{tom} datasets and evaluated their performance and reasoning processes across multiple benchmarks.

Our results demonstrate that while RL effectively improves accuracy on \ac{tom} tasks across model scales, its impact on the quality of reasoning is scale-dependent. In larger models, RL fostered structured, interpretable belief tracking that generalized to higher-order reasoning and transferred knowledge effectively to another model. In contrast, smaller models ($\leq$3B) exhibited reasoning collapse, achieving high accuracy via drastically shortened, less interpretable outputs. Furthermore, we found that simple \ac{sft} alone yields surprisingly competitive and generalizable performance on current benchmarks.

These findings underscore a critical challenge in the pursuit of artificial social intelligence: achieving high performance metrics on existing benchmarks does not guarantee robust, interpretable reasoning, particularly in smaller models. The discrepancy between \ac{tom} accuracy and mind reasoning quality revealed by our analysis suggests that current evaluation methods may be insufficient to probe the underlying cognitive processes models employ.

Moving forward, it is crucial to:
\begin{itemize}
    \item Develop training methodologies, potentially involving more sophisticated reward designs or architectural innovations, that encourage the learning of interpretable \ac{tom} reasoning.
    \item Create \ac{tom} benchmarks that go beyond measuring final answer correctness to explicitly evaluate the quality, consistency, and interpretability of models' belief attribution and reasoning processes.
\end{itemize}

By focusing on both performance and the underlying mechanisms, we can pave the way towards building LLMs that not only perform well on social tasks but also genuinely understand and reason about the minds of others, fostering more natural and trustworthy human-AI interaction.
\newpage
\bibliographystyle{abbrvnat}
\bibliography{references}

\clearpage

\appendix
\section{Datasets Description}
\paragraph{Hi-ToM dataset}
The Hi-ToM  dataset\citep{heHITOMBenchmarkEvaluating2023} is specifically designed to systematically evaluate LLMs' \ac{tom} reasoning across varying orders of complexity. It comprises multi-agent stories set within an object-finding game, where five agents interact in one or more rooms containing objects and containers. The dataset is automatically generated using scripts. The synthetic generation process involves fitting randomly chosen story components (rooms, objects, containers, agents) into pre-defined chapter templates, which can include multiple actions like object movement and agent communication. Each story is coupled with five progressively higher-order questions, as exemplified by: "Where is O really?" (0th-order), "Where does A1 think O is?" (1st-order), and extending up to 4th-order nested beliefs.

\paragraph{Explore-ToM dataset}
This dataset\citep{sclarExploreTheoryMind2024} employs an A*-guided adversarial generation process to create complex stories that challenge models' \ac{tom} reasoning. Instead of sampling different elements independently, this dataset uses LLM to sample plausible story contexts (characters, locations, objects, etc.). Subsequently, an A* search algorithm explores sequences of actions defined in a domain-specific language, such as object movements, to construct story structures that are particularly challenging for target LLMs. Each story is paired with questions probing first-order, second-order, and state-tracking abilities. These structured stories are optionally rewritten with language models to produce naturalistic narratives. The whole dataset consists of 13309 questions with 619 unique raw stories.

\paragraph{ToMi dataset}
The ToMi dataset \citep{le-etal-2019-revisiting} is a ToM benchmark generated by scripts. The current ToMi dataset consists of 5994 questions covering unexpected transfers as well as first- and second-order false belief tasks.

\begin{table}[h!]
\centering
\caption{Examples from evaluation datasets.}
\small
\begin{tabularx}{\textwidth}{p{0.14\textwidth} p{0.56\textwidth} p{0.25\textwidth}}
\toprule
\textbf{Dataset} & \textbf{Story} & \textbf{Question \& Answer} \\
\midrule
Hi-ToM & Olivia, Chloe, Oliver, Lily and Avery entered the playroom. The tangerine is in the blue\_crate. Lily likes the blue\_treasure\_chest. Olivia moved the tangerine to the blue\_pantry. Olivia likes the red\_pantry. Olivia exited the playroom. Chloe made no movements and stayed in the playroom for 1 minute. Avery lost his watch. Chloe exited the playroom. Oliver made no movements and stayed in the playroom for 1 minute. Oliver exited the playroom. Lily moved the tangerine to the red\_bottle. Lily exited the playroom. Avery moved the tangerine to the blue\_crate. Avery exited the playroom. Olivia, Chloe, Oliver, Lily and Avery entered the waiting\_room. Olivia, Lily and Oliver entered the playroom. The tangerine is in the blue\_crate. Olivia moved the tangerine to the blue\_pantry. Olivia exited the playroom. Lily made no movements and stayed in the playroom for 1 minute. Lily exited the playroom. Oliver made no movements and stayed in the playroom for 1 minute. Oliver exited the playroom. Olivia, Lily and Oliver entered the waiting\_room. Olivia saw a monkey. Olivia, Oliver, Lily and Avery entered the playroom. The tangerine is in the blue\_pantry. Olivia made no movements and stayed in the playroom for 1 minute. Olivia exited the playroom. Oliver lost his phone. Oliver moved the tangerine to the red\_drawer. Oliver exited the playroom. Lily moved the tangerine to the green\_suitcase. Lily exited the playroom. Avery made no movements and stayed in the playroom for 1 minute. Avery exited the playroom. Oliver lost his gloves. Olivia, Oliver, Lily and Avery entered the waiting\_room. & {\raggedright Question: Where does Oliver think Chloe thinks Lily thinks Avery thinks the tangerine is?\\Answer: blue\_pantry} \\
\addlinespace
ToMi & Amelia entered the hall. Mila loves the plum Hunter loves the tomato Mila entered the lounge. Hunter entered the hall. The jeans is in the container. Amelia moved the jeans to the treasure\_chest. Hunter exited the hall. Amelia exited the hall. Hunter entered the hall. & {\raggedright Question: Where does Amelia think that Hunter searches for the jeans?\\Answer: treasure\_chest} \\
\addlinespace
{\raggedright ExploreTom\\(structure version)} & Brody entered the back room of the thrift store. Brody moved the vintage typewriter to the cardboard box, which is also located in the back room of the thrift store. While this action was happening, Lucas witnessed this action in secret (and only this action). Evelyn entered the back room of the thrift store. Lucas entered the back room of the thrift store. Evelyn moved the vintage typewriter to the plastic storage bin, which is also located in the back room of the thrift store. & {\raggedright Question: In which container will Evelyn search for the vintage typewriter? \\ Answer: lastic storage bin} \\
\addlinespace
{\raggedright ExploreTom\\(infilled version)} & The back room of the thrift store was quiet and dimly lit, with cardboard boxes stacked against the walls and morning sunlight barely peeking through the grimy high window. The air was thick with the smell of old fabrics and forgotten items, and the silence was almost palpable, broken only by the faint hum of the old store. As Brody entered the back room, the soft creak of the door seemed to amplify in the stillness, a solitary sound that broke the morning calm. In a move that didn't escape Lucas's watchful gaze, the vintage typewriter glided off its perch, landing softly within the corrugated walls of a cardboard box in the corner. With a gentle creak, the back room door swung open for Evelyn, who swept in with a purposeful stride, closely followed by Lucas. The back room was once again quiet, the only sign of activity being the soft clinking of the plastic storage bin's lid as Evelyn securely placed the vintage typewriter inside, her task complete.& {\raggedright Question: In which container will Evelyn search for the vintage typewriter?\\Answer: plastic storage bin} \\
\bottomrule
\end{tabularx}
\label{tab:datasets}
\end{table}
\clearpage

\section{Prompt design}
\subsection{System prompts}
\begin{tcolorbox}[title=System prompt for RL training, mypromptbox]
You are a helpful assistant. The assistant first thinks about the reasoning process in the mind and then provides the user with the answer. The reasoning process and answer are enclosed within <think> </think> and <answer> </answer> tags, respectively, i.e., <think> reasoning process here </think><answer> answer here </answer>. Now the user asks you to solve a \ac{tom} reasoning problem. After thinking, when you finally reach a conclusion, clearly state your answer within <answer> </answer> tags.\textbackslash nNote: You should assume the following.\textbackslash n(1) An agent witnesses everything and every movement before exiting a room.\textbackslash n(2) An agent A can infer another agent B's mental state only if A and B have been in the same room, or have private or public interactions.
\end{tcolorbox}\label{tcolorbox:sys_prompt}
\vspace{3em}

\begin{tcolorbox}[float*, title=CoT system prompt for baseline evaluation, label={tcolorbox:cot_prompt}, colback=gray!5, colframe=gray!80, fonttitle=\bfseries]Read the following story and answer the question. Think step-by-step. Provide the thinking first, and then the answer. Answer in the following JSON format:\\
\{\\
"thinking": "step by step thinking",\\
"answer": "answer text"\\
\}
\end{tcolorbox}
\vspace{3em}
\paragraph{Prompts for knowledge transfer} For knowledge transfer experiment, we used the CoT system prompt and simply added the models' RL thinking content to user message: ``\textit{Story: \{story\} \textbackslash n Question: \{question}\} \textbackslash n <think>\{thinking\}</think>''.
For baseline models, the reasoning content is extracted from the "thinking" field in the JSON response. For RL models, it corresponds to the text enclosed within the <think>...</think> tags. We evaluated both the original reasoning content and a version that removed the final summarizing statement.

\paragraph{Assumptions in prompts} In the original HI-TOM\cite{heHITOMBenchmarkEvaluating2023} dataset, they attached the following assumption list to each story:
\begin{verbatim}
Note: You should assume the following.
(1) An agent witnesses everything and every movement before exiting a room.
(2) An agent A can infer another agent B’s mental state only if A and B have 
been in the same room, or have private or public interactions.
(3) Note that every agent tends to lie. What a character tells others doesn’t
affect their actual belief. An agent tends to trust an agent that exited the
room later than themselves. The exit order is known to all agents.
(4) Agents in private communications know that others won’t hear them, but
they know that anyone can hear any public claims.
\end{verbatim}
\textbf{We only kept the first 2 assumptions} for RL training. When evaluating the baseline models, we found that these assumptions did not improve performance, and in some cases even degraded it. Therefore, we excluded them during evaluation.

\clearpage
\subsection{Logical coherence prompts}
\begin{tcolorbox}[float*, title=Thinking quality evaluation prompts, label={tcolorbox:think_prompt}, colback=gray!5, colframe=gray!80, fonttitle=\bfseries]
You are an expert in evaluating Theory of Mind reasoning. Your task is to assess a student's explanation for a complex Theory of Mind question.\\
\\
The evaluation will proceed as follows:\\
1. You will first read a Theory of Mind question and the correct answer.\\
2. Then, you will be shown a student's thinking process in response to the question.\\
3. Please evaluate the student's thinking process based on the following criteria:\\

- **Logical Coherence (0–10 points)**: Does the reasoning make sense as a whole? Evaluate whether the thinking is internally consistent, logically structured, and meaningfully sequenced.\\
  - 0 – Completely incoherent: ideas are contradictory, disorganized, or temporally scrambled.\\
  - 5 – Partially coherent: some valid reasoning exists, but the steps are confusing, redundant, or lack a clear flow.\\
  - 10 – Fully coherent: reasoning is orderly, step-by-step, and clearly follows the timeline and causal structure of the events and beliefs.\\
- **Evaluation**: Write a short explanation (1–2 sentences) justifying your judgment, pointing out any key strengths or flaws.\\

Return your evaluation in the following JSON format:\\
{{"LogicalCoherence": 0–10, "Evaluation": "A short explanation"}}\\
\\
Here is the question and the correct answer:\\
Question: \{question\}\\
Answer: \{answer\}\\
\\
Here is the student's thinking process:\\
\{thinking\}\\
\end{tcolorbox}
\vspace{3em}

\newpage
\subsection{Factual alignment prompts}
\begin{tcolorbox}[float*, title=Factual alignment evaluation prompts, label={tcolorbox:think_prompt2}, colback=gray!5, colframe=gray!80, fonttitle=\bfseries]
You are an expert in evaluating Theory of Mind reasoning. Your task is to assess a student's explanation for a complex Theory of Mind question based on a given story.\\
\\
The evaluation will proceed as follows:\\
1. You will first read a story, a Theory of Mind question based on that story, and the correct answer.\\
2. Then, you will be shown a student's thinking process in response to the question.\\
3. Please evaluate the student's explanation based on the following criteria:\\
\\
- **Factual Alignment (0–10 points)**: Does the reasoning accurately reflect the facts in the story? Evaluate whether the steps in the student’s thinking are grounded in the actual events, character actions, and timelines described.\\
  - 0 – Major factual errors or hallucinations; reasoning contradicts the story.\\
  - 5 – Some facts are correct, but key details are omitted, misremembered, or inaccurately applied.\\
  - 10 – All relevant details are accurate and clearly support the reasoning.\\
- **Evaluation**: Write a short explanation (1–2 sentences) justifying your judgment, pointing out any key strengths or flaws.\\
\\
Return your evaluation in the following JSON format:\\
{{"FactualScore": 0–10, "Evaluation": "A short explanation"}}\\
\\
Here is the story and the correct answer:\\
\{story\}\\
Question: \{question\}\\
Answer: \{answer\}\\
\\
Here is the student's thinking process:\\
\{thinking\}\\
\end{tcolorbox}
\vspace{3em}

\newpage
\section{Model responses example}
\subsection{4th-Order reasoning example}
\begin{tcolorbox}[title=4th-Order Reasoning Process Example, mypromptbox, label={tcolorbox:response_example}]
\textbf{Story \& Question: } \\
Ella, Mila, Benjamin, Gracie and William entered the TV\_room. The tomato is in the red\_envelope. Ella made no movements and stayed in the TV\_room for 1 minute. Ella lost his phone. Ella exited the TV\_room. Mila moved the tomato to the blue\_bathtub. Mila exited the TV\_room. Benjamin made no movements and stayed in the TV\_room for 1 minute. Benjamin exited the TV\_room. Gracie moved the tomato to the red\_bucket. Gracie exited the TV\_ room. William moved the tomato to the red\_envelope. William exited the TV\_room. Benjamin saw a cat. Ella, Mila, Benjamin, Gracie and William entered the waiting\_room. Benjamin privately told Ella that the tomato is in the blue\_bathtub now.
\\
\\
Where does Gracie think William thinks Benjamin thinks Ella thinks the tomato is?
\\
\\
\textbf{7B-1M RL Model Response: } \\
<think>\\
Based on the story, initially the tomato is in red\_envelope. Ella exits before anything is moved, so she thinks tomato is in red\_envelope. Mila moves tomato to blue\_bathtub and exits. Benjamin sees tomato in blue\_bathtub and exits. Gracie moves tomato to red\_bucket and exits. William moves tomato to red\_envelope and exits. Ella leaves before tomato is moved, so she still thinks tomato is in red\_envelope. Mila exits before telling Ella, so this doesn't affect Ella's belief. \boldgreen{Benjamin privately tells Ella tomato is in blue\_bathtub, but this is a private interaction and doesn't change the initial beliefs of others}. \boldgreen{When Gracie moves tomato, it doesn't affect Benjamin's or Ella's belief. When William moves tomato, it also doesn't change the beliefs of the ones who left before that}. So, when Ella leaves initially, she thinks tomato is in red\_envelope. Mila leaves before telling Ella, and Benjamin leaves before Gracie moves the tomato. So Gracie thinks William thinks Benjamin thinks Ella thinks the tomato is in red\_envelope. \\
</think>\\
<answer>\\
Gracie thinks William thinks Benjamin thinks Ella thinks the tomato is in the red\_envelope.\\
</answer>\\
\\
\textbf{3B RL Model Response: } \\
<think>Initially, Ella made no movements and stayed in the TV\_room for 1 minute with the tomato in the red\_envelope. \boldred{Since Ella last saw the tomato in the red\_envelope, when Benjamin last saw the tomato, it would still be in the red\_envelope}. Therefore, Gracie would think William still thinks Ella thinks the tomato is in the red\_envelope.
</think>\\
<answer>red\_envelope</answer>\\
\\
\textbf{1.5B RL Model Response: } \\
<think>\boldred{Gracie thinks Ella thinks the tomato is in the blue\_bathtub}. </think>\\
<answer> blue\_bathtub </answer>\\
\\
\textbf{0.5B RL Model Response: } \\
<think> \boldred{reasoning process here} </think>\\
<answer> Gracie thinks William thinks Benjamin thinks Ella thinks the tomato is in the red\_envelope </answer>\\
\end{tcolorbox}

\newpage
\subsection{Wrong thinking processes examples}
\begin{tcolorbox}[title=ToMi Reasoning Process Example with Wrong Thinking, mypromptbox, label={tcolorbox:response_example_wrong}]
\textbf{Story \& Question: } \\
Story: Emily entered the dining\_room. Ava entered the dining\_room. The pumpkin is in the basket. Ava exited the dining\_room. Ava entered the study. Avery dislikes the strawberry. Emily moved the pumpkin to the bottle.\\
\\
Question:  Where does Emily think that Ava searches for the pumpkin?\\

\textbf{7B-1M RL Model Response} (wrong answer and thinking): \\
<think>\\
 Based on the story, initially the pumpkin is in the basket. Ava exits the dining room before entering the study, and \boldred{Emily moves the pumpkin to the bottle. After this, Ava enters the study}. Emily last saw the pumpkin when she put it in the bottle. Therefore, Emily thinks that Ava searches for the pumpkin in the bottle. </think>\\
<answer>Emily thinks that Ava searches for the pumpkin in the bottle. </answer>\\
\\
\textbf{3B RL Model Response} (correct answer with wrong thinking): \\
<think>Initially, \boldred{since Ava last had the pumpkin in the basket and there are no other mentions of its relocation, Emily would likely think Ava searches for the pumpkin in the basket}. Therefore, Emily thinks Ava searches for the pumpkin in the basket.
</think>\\
<answer>basket</answer>\\

\textbf{DeepSeek-v3 Response} (correct answer and thinking): \\
"thinking": "To answer the question, let's break down the story step by step:\\

1. Initial Positions and Actions:\\
   - Emily and Ava both enter the dining room.\\
   - The pumpkin is initially in the basket in the dining room.\\
   - Ava exits the dining room and enters the study.\\
   - Emily moves the pumpkin from the basket to the bottle (location not specified, but likely still in the dining room since Emily hasn't left).\\

2. Key Observations:\\
   - Ava is no longer in the dining room when Emily moves the pumpkin; she is in the study.\\
   - There is no information suggesting that Ava knows about the pumpkin being moved to the bottle.\\
   - The question asks where Emily thinks Ava would search for the pumpkin, not where Ava actually searches.\\

3. Emily's Perspective:\\
   - Emily knows the pumpkin was originally in the basket in the dining room.\\
   - Emily moved the pumpkin to the bottle, but Ava was not present to witness this change.\\
   - From Emily's perspective, Ava's last knowledge of the pumpkin's location was in the basket in the dining room.\\

4. Conclusion:\\
   - Since Ava was not present when the pumpkin was moved, Emily would assume Ava still believes the pumpkin is in the basket in the dining room.\\

Thus, Emily thinks Ava would search for the pumpkin in the basket in the dining room.",\\

"answer": "Emily thinks that Ava searches for the pumpkin in the basket in the dining room."\\
\end{tcolorbox}

\newpage
\subsection{Example of thinking in structured and infilled stories of ExploreToM}
\begin{tcolorbox}[title=Example of thinking in structure and infilled stories of ExploreToM, mypromptbox, label={tcolorbox:response_example_exploretom}]
\textbf{Structured Story: } \\
Story: Abigail entered the backstage tent. Abigail moved the portable led flashlight to the metal toolbox, which is also located in the backstage tent. Abigail moved the portable led flashlight to the black nylon backpack, which is also located in the backstage tent. While this action was happening, Amelia witnessed this action in secret (and only this action). Kayla entered the backstage tent. Amelia entered the backstage tent. Amelia moved the portable led flashlight to the metal toolbox, which is also located in the backstage tent.\\
\\
Question:  In which room does Abigail think that Kayla will search for the black nylon backpack?\\

\textbf{7B-1M RL Model Response}: \\
<think>\\
 Based on the story, Abigail initially placed the portable LED flashlight in a metal toolbox and then moved it to a black nylon backpack, both of which were in the backstage tent. Later, Amelia moves the flashlight back to the metal toolbox, but this doesn't affect the location of the black nylon backpack. When Abigail leaves, the black nylon backpack is still in the backstage tent. Abigail leaves before Kayla enters. Abigail does not leave the backstage tent. Thus, when Abigail leaves, she thinks the black nylon backpack is in the backstage tent. Therefore, Abigail thinks that Kayla will search for the black nylon backpack in the backstage tent. </think>\\
<answer>Abigail thinks that Kayla will search for the black nylon backpack in the backstage tent. </answer>\\
\\
\textbf{Infilled Story: } \\
Story: The music festival was in full swing, with vibrant lights and flags scattered across the rolling hills of the outdoor venue. The backstage area, a hub of behind-the-scenes activity, was a flurry of tents and equipment, surrounded by the quiet woods that bordered the festival grounds. The sudden change from the bright sunlight to the dim interior of the tent was refreshing for Abigail, the cool shade a welcome respite from the relentless sun beating down on the festival grounds; as she moved deeper into the tent, she began checking her mental list of tasks. With every passing minute, the energy inside the backstage area grew more intense, and the production's intricate machinery whirred into motion. Abigail swiftly tucked the portable LED flashlight into different places in the backstage tent, moving it from the metal toolbox to the black nylon backpack as she fine-tuned the setup, and Amelia happened to observe these quiet adjustments from a discreet distance. As Amelia stepped inside the backstage tent, the sounds of the festival outside receded into the distance, replaced by the hum of generators and quiet chatter of crew members, while her gaze swept the space, taking in every detail. Meanwhile, Kayla's eyes adjusted to the dim light, and she made her way deeper into the tent, her senses alert to the unfamiliar surroundings. As she absorbed every element of the bustling tent, Amelia leaned over to slip the flashlight into the metal toolbox, where it belonged.\\
\\
Question:  In which room does Abigail think that Kayla will search for the black nylon backpack?\\

\textbf{7B-1M RL Model Response}: \\
<think>\\
 Based on the story, Abigail initially places the portable LED flashlight in a metal toolbox and a black nylon backpack in the backstage tent. Later, Amelia slips the flashlight into the metal toolbox, and there is no indication that anything is moved outside the tent. When Kayla makes her way deeper into the tent, she is still inside the backstage tent. Therefore, Abigail would think that Kayla will search for the black nylon backpack in the backstage tent. </think>\\
<answer>Abigail thinks that Kayla will search for the black nylon backpack in the backstage tent.</answer>\\
\end{tcolorbox}

\clearpage
\section{Appendix tables}
\begin{table}[ht]
\centering
\caption{Comparison of GSM8K performance between baseline and RL-trained Qwen models. }
\begin{tabular}{lcc}
\toprule
Model & Baseline & RL-based \\
\midrule
Qwen2.5-0.5B-Instruct & 47.99\% & 45.49\% \\
Qwen2.5-1.5B-Instruct & 70.28\% & 67.48\% \\
Qwen2.5-3B-Instruct & 86.73\% & 86.20\% \\
Qwen2.5-7B-Instruct & 92.04\% & 91.58\% \\
Qwen2.5-7B-Instruct-1M & 91.13\% & 90.83\% \\
\bottomrule
\end{tabular}
\label{tab:gsm8k_results}
\end{table}
\vspace{3em}

\begin{table}[htbp]
\centering
\caption{Evaluation scores of different models' thinking quality.}
\vspace{1mm}
\begin{tabular}{lccc}
\toprule
Model & Logical Coherence & Factual Score & Mean Thinking Score \\
\midrule
DeepSeek-v3                & \textbf{8.608} & \textbf{7.920} & \textbf{0.826} \\
GPT-4o                     & 7.565 & 6.982 & 0.727 \\
GPT-4o-mini                & 5.373 & 5.195 & 0.528 \\
Qwen2.5-3B-Instruct        & 4.700 & 3.752 & 0.423 \\
Qwen2.5-7B-Instruct-1M     & 5.582 & 4.712 & 0.515 \\
\midrule
Qwen2.5-3B-Instruct (RL)   & 6.243 & 5.715 & 0.598 \\
Qwen2.5-7B-Instruct-1M (RL)& \textbf{6.868} & \textbf{7.468} & \textbf{0.717} \\
\bottomrule
\end{tabular}
\label{tab:model-eval}
\end{table}
\vspace{3em}

\begin{table}[thbt]
\centering
\caption{SFT Performance of Qwen2.5 on ToM Benchmarks with Varied Training Data}
\begin{tabular}{lccccc}
\toprule
Model & Hi-ToM & \makecell{4th-order\\ToM} & ToMi & ExploreToM & \makecell{ExploreToM\\(Infilled)} \\
\midrule
Qwen2.5-0.5B-Instruct (800)  & 70.83\% & 80.33\% & 78.35\% & \textbf{92.87}\% & 72.05\%\\
Qwen2.5-1.5B-Instruct (800)  & 69.67\% & 71.83\% & 78.73\% & 89.77\% & 69.70\%\\
Qwen2.5-3B-Instruct (800)    & 69.00\% & 69.33\% & 77.08\% & 84.24\% & 72.23\%\\
Qwen2.5-7B-Instruct (800)    & \textbf{81.67}\% & \textbf{88.83}\% & 76.86\% & 90.06\% & 72.14\%\\
Qwen2.5-7B-Instruct-1M (800) & 80.50\% & 86.33\% & \textbf{81.01}\% & 92.03\% & \textbf{75.70}\%\\
\midrule
Qwen2.5-0.5B-Instruct (1600)  & 70.83\% & 83.33\% & 76.33\% & 90.81\% & 70.83\% \\
Qwen2.5-1.5B-Instruct (1600)  & 63.33\% & 61.83\% & 76.39\% & 88.84\% & 72.89\%\\
Qwen2.5-3B-Instruct (1600)    & 82.50\% & 88.33\% & 77.76\% & 93.34\% & 74.86\%\\
Qwen2.5-7B-Instruct (1600)    & \textbf{89.83}\% & \textbf{93.00}\% & \textbf{79.26}\% & \textbf{95.22}\% & \textbf{75.89}\%\\
Qwen2.5-7B-Instruct-1M (1600) & 68.00\% & 64.67\% & 77.01\% & 93.15\% & 75.61\% \\
\midrule
Qwen2.5-0.5B-Instruct (3200) & 81.00\% & 88.17\% & 77.79\% & 89.68\% & 69.89\% \\
Qwen2.5-1.5B-Instruct (3200) & 80.50\% & 86.17\% & 76.33\% & 93.53\% & 74.67\% \\
Qwen2.5-3B-Instruct (3200) & 87.00\% & 92.67\% & 79.55\% & 95.78\% & 74.95\% \\
Qwen2.5-7B-Instruct (3200) & \textbf{87.33\%} & \textbf{94.00\%} & 80.85\% & \textbf{95.97\%} & \textbf{77.95\%} \\
Qwen2.5-7B-Instruct-1M (3200) & 86.50\% & 93.67\% & \textbf{81.10\%} & 95.12\% & 75.61\% \\
\bottomrule
\end{tabular}
\label{tab:sft_results}
\end{table}

\clearpage
\end{document}